# Boosting Cross-Quality Face Verification using Blind Face Restoration

BENGHERABI Messaoud
*Centre de Développement et des Technologies Avancées (CDTA)*
Algiers, Algeria
mbengherabi@cdta.dz

LAIB Douaa
*Ecole Nationale Superieure d'Informatique (ESI)*
Algiers, Algeria
hd_laib@esi.dz

LASNAMI Fella Souhila
*Ecole Nationale Superieure d'Informatique (ESI)*
Algiers, Algeria
if_lasnami@esi.dz

BOUSSAHA Ryma
*Ecole Nationale Superieure d'Informatique (ESI)*
Algiers, Algeria
r_boussaha@esi.dz

*Abstract*—In recent years, various Blind Face Restoration (BFR) techniques were developed. These techniques transform low quality faces suffering from multiple degradations to more realistic and natural face images with high perceptual quality. However, it is crucial for the task of face verification to not only enhance the perceptual quality of the low quality images but also to improve the biometric- utility face quality metrics. Furthermore, preserving the valuable identity information is of great importance. In this paper, we investigate the impact of applying three state-of-the-art blind face restoration techniques namely, GFP-GAN, GPEN and SGPN on the performance of face verification system under very challenging environment characterized by very low quality images. Extensive experimental results on the recently proposed cross-quality LFW database using three state-of the -art deep face recognition models demonstrate the effectiveness of GFP-GAN in boosting significantly the face verification accuracy.

## I. INTRODUCTION

Recent advances in deep learning techniques and the availability of very large-scale datasets have resulted in drastic performance improvement in facial recognition systems [1]–[3]. This progress makes the Face Recognition Technology (FRT) a prominent tool for identity verification and identification in various applications ranging from simple access control to intelligent video surveillance and advanced smart safe city applications [4]. However, the near-perfect accuracy surpassing 99.8% obtained on the Labeled Faces in the Wild (LFW) database [5] is not generalizable to more challenging realistic conditions, especially outdoor distant face recognition which still presents one of the main challenges in video surveillance and face-related forensic systems. Real-world scenarios pose unavoidable challenges. In these scenarios, images frequently suffer from various distortions such as noise [6], blur [7], and low resolution [8]. These degradations significantly hinder the ability of face recognition systems to accurately identify and distinguish facial features, resulting in a notable decline in their overall performance [9]. To address these aforementioned challenges, a range of solutions has been proposed where face enhancement techniques play a pivotal role.

Face restoration aims to restore a high-quality facial image from its low-quality counterpart by eliminating both known and unknown degradations present in such images [10]. Traditionally, known degradations can be addressed through non-blind restoration techniques, which have been effective in targeting specific types of image distortions [11]. These approaches include deblurring techniques to remove blur [12], [13], denoising techniques to reduce noise [14], super-resolution techniques to enhance resolution [15], [16], and compression artifact removal techniques [17].

While these classical techniques yield reasonably reliable results, real-world captured images often exhibit complex and heterogeneous degradations, making the accurate estimation of the type of degradation a very difficult task. With the emergence of deep learning and the rapid advancements in Convolutional Neural Networks CNNs [18], [19] and deep Generative Adversarial Networks GANs [20] [21], blind face restoration techniques [10] have emerged to address cases where the degradation type is unknown or multiple types of degradations coexist within the same image. These techniques can be classified into two main categories: non-priors restoration techniques and priors restoration techniques [11]. Non-priors restoration techniques focus on restoring degraded facial images without relying on any prior information about the degradation process [22], [23]. In contrast, priors' restoration techniques leverage prior knowledge to guide the restoration process. These priors can be in the form of reference information [18], [19], geometric constraints [24], [25], or generative priors [20], [26].

Previous studies primarily focused on evaluating the effectiveness of these methods in improving the perceptual quality of enhanced images using image quality assessment IQA [27] and face quality assessment FQA [28] metrics. However, an essential aspect that has often been overlooked is the preservation of identity information within the enhanced images and its impact on face recognition performance. In this study, we aim to bridge this gap by evaluating the influence of face restoration techniques in terms of their practical utility on the accuracy and reliability of face recognition systems. Our objective is to assess the extent to which these techniques not only enhance the visual quality of degraded facial images but also preserve the crucial identity information necessary for accurate face recognition. To achieve these objectives, we conducted a rigorous evaluation process that involved comparing the performance of state-of-the-art face recognition models, including AdaFace [29], MagFace [30], and ArcFace [31]. Our evaluation encompassed both the original low-quality images from the XQLFW [32] dataset and the enhanced images generated using blind face restoration techniques, including the Generative Facial Prior GAN (GFP-GAN) [20], the GAN Prior Embedded Network (GPEN) [21], and the Shape and Generative Prior integrated Network (SGPN) [26]. By evaluating the performance of face recognition models on restored images, we not only explore the potential of face restoration techniques in improving face recognition accuracy but also shed light on the extent to which these techniques preserve the identity information crucial for accurate recognition. The main contributions of this work are summarized as follows: First, we investigate the impact of applying Blind Face Restoration techniques on the face utility biometric quality metric [33] represented in this study by the amplitude of MagFace embedding vector [30]. Second, we quantify the identity preserving capability of each restoration technique by computing the similarity index obtained from the statistics of the cosine similarity between the original LFW images and the restored XQLFW images. Third, we investigate the impact of the three Blind Face Restoration techniques on the performance of three state-of-the-art face recognition models. By comparing their performance on the restored images, we gained valuable insights and shed light that BFR preprocessing techniques can significantly boost the performance of face verification systems under multiple image degradations if and only if the resulting restored faces possess both higher biometric quality and higher identity preserving similarity index. The rest of this paper is organized into three sections. The second section provides an overview of the three blind face restoration techniques used in our study. We discuss their principles, advantages, and limitations. The third section presents our proposal for a Facial Verification System Architecture specifically designed to operate in low-quality environments. In the fourth section, we present the experimental results obtained from our investigations. We analyze and discuss the findings, shedding light on the influence of these techniques on face verification accuracy and the overall biometric utility of face recognition. Finally, we conclude the paper with a comprehensive discussion and perspectives on the implications of our research.

## II. BLIND FACE RESTORATION VIA GAN PRIORS

Blind face restoration techniques based on generative priors, including GFPGAN, GPEN, and SGPN, employ the strength of Generative Adversarial Networks (GANs) [34] to generate realistic representations of the original image. By filling in the missing or altered information in damaged images, these GAN-generated representations capture intricate details of facial geometry, regional textures, and accurate color information. These representations then serve as valuable references for restoring degraded images, resulting in improved visual quality and enhanced facial appearance [20], [26], [21].

GFP-GAN [20] is an architecture capable of restoring facial details and enhancing image colors in a single pass. It combines a U-Net [35] degradation elimination module and a pre-trained StyleGAN2 [36]. The restoration process is guided by four cost functions: adversarial loss for realistic textures, reconstruction loss for preserving fine details and overall quality, Facial Component Loss for enhancing specific facial regions, and identity preserving loss for maintaining accurate identities. This guidance ensures that the restored images accurately represent the original identities. However, GFP-GAN may face challenges when dealing with images exhibiting extreme poses. GPEN [21] unlike conventional methods that aim to learn a direct mapping from low-quality (LQ) input images to high-quality (HQ) images, adopts a two-step approach: training a StyleGAN [37] on high-quality images to capture desired visual characteristics and generate realistic images, and using a decoder-encoder architecture to reconstruct global face structure and local facial details. It employs three cost functions: content loss to preserve fine features and color information, feature matching loss to enhance realism, and adversarial loss for more vivid details. This approach enables GPEN to generate high-quality images with rich details and reduced smoothness. SGPN [26] aims to restore faithfully both the shape and detail of the face. It consists of two main modules: the shape restoration module, which reconstructs facial geometry using 3DMM [38], and the Shape and Generative prior Integration module, which generates a high-quality image by combining reconstructed shape and texture information. SGPN shows promising performance, especially in extreme exposure images. However, it has been observed that SGPN may not fully preserve the identity in the restored images.

## III. METHODOLOGY

It is a common practice in modern face verification systems based on deep learning to start with face detection and alignment. These necessary steps are followed immediately by the embedding operation using one of stare-of-the-art pretrained models. The objective is to extract low dimensional and highly discriminative feature vectors. Generally, a simple cosine similarity scoring is used for matching. However, low-quality images suffering from distortions present less distinguishable features. As a result, face recognition models encounter difficulties when extracting feature vectors from these images, leading to a decrease in face verification accuracy [39]. To overcome this challenge, we propose the inclusion of a face restoration module to the conventional verification pipeline. The new architecture that incorporates blind face restoration is depicted in Fig. 1. By incorporating this face restoration module in the early stage of the face verification process, we ensure that the images, possessing improved quality, exhibit more distinguishable features.

## IV. EXPERIMENTS

### A. Datasets

In our evaluation process, we employed two different datasets. For the initial phase of investigating the impact of the studied blind face restoration techniques on the verification performance of state of the art FR systems, we utilized the widely recognized LFW (Labelled Faces in the Wild) dataset [5] under the restricted protocol. In the subsequent phases, specifically for investigating the effectiveness of face restoration techniques, we employed the XQLFW (Cross-Quality Faces in the Wild) dataset which is a variant of LFW with synthetic degradations [32]. This dataset was designed to address the challenges of facial recognition in degraded image conditions. It includes a range of images with varying levels of degradation, such as low quality, low resolution, and low illumination. Fig. 2 shows a sample of positive and negative pairs from the two datasets.

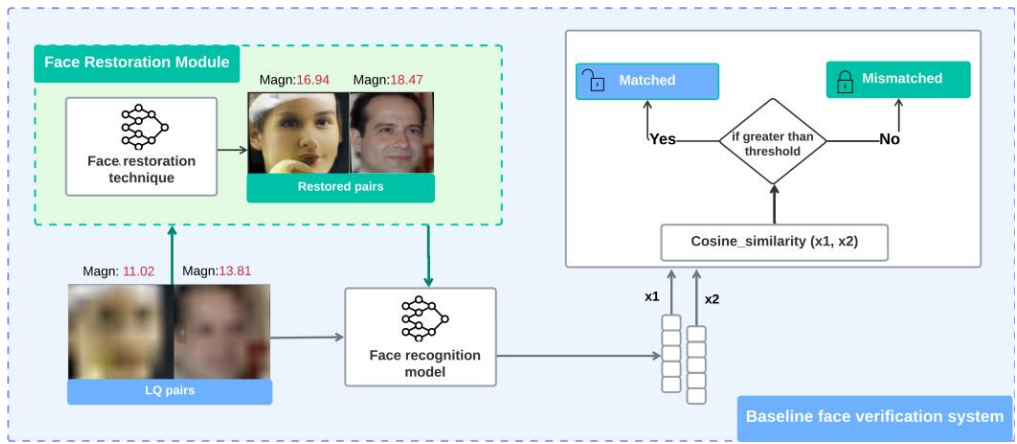

Fig. 1. Architecture of the face verification pipeline incorporating blind face restoration. *Magn* indicates the corresponding MagFace quality metric of the image.

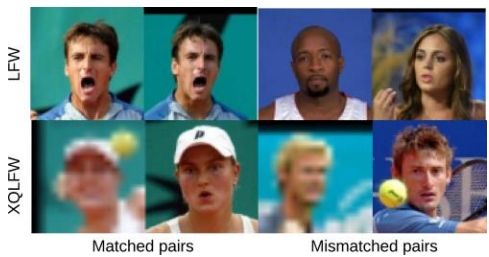

Fig. 2. Sample of positive and negative pairs from the LFW and the XQLFW datasets.

### B. Face Recognition Techniques

In this evaluation process, we utilized three state-of-the-art face recognition systems: ArcFace, MagFace[1], and AdaFace [1]. ArcFace [31] is a face recognition technique that improves upon the traditional Softmax loss by introducing an additive margin loss. By considering the angular margin between classes, ArcFace enhances the model's ability to discriminate and classify different identities, resulting in high accuracy in face recognition tasks. This technique has laid a solid foundation for advancements in the field. MagFace [30] builds upon ArcFace by refining the margin concept and incorporating additional considerations. It integrates magnitude considerations into the learning process, enabling the model to capture subtle variations in facial features caused by lighting conditions, facial expressions, and other factors. This enhancement allows MagFace to outperform previous approaches and achieve superior performance in face recognition tasks. AdaFace [29], also inspired by ArcFace, introduces an adaptive margin mechanism that adjusts the margin dynamically during training based on the quality of the input image. By considering image quality, AdaFace can handle variations in image clarity and other factors that affect the quality of facial features, such as occlusion [40].

### C. Implementation details

In our comprehensive evaluation of face restoration techniques, we conducted experiments in different image quality scenarios: on high-quality images, low-quality images, and re- stored images. For the first experiment on high-quality images using LFW dataset, face detection and alignment are accomplished using MTCNN [41]This experiment served as a baseline for the rest of the evaluation process. Similar procedure is executed when evaluating the verification performance on XQLFW and their restored versions. For blind face restoration implementation the **GFP-GAN v1.3.0**[2], **GPEN-BFR-512**[3], and **SGPN**[4] pretrained models are employed in this study.

### D. Experimental Results

*1) Preliminary Visual Inspection:* As depicted in Fig. 3, the effectiveness of the three restoration techniques (GFP-GAN, GPEN, and SGPN) varies based on the degree of image degradation. Mildly degraded images are successfully restored quality by all the three techniques with a high perceptual image quality. However, as distortion increases, SGPN encounters difficulties in restoring facial components, particularly the eyes, while GPEN and GFP-GAN perform relatively better but with reduced naturalism and fidelity compared to the original images. This is evident in cases where non-existent glasses or changes in eye color are introduced. These artifacts yield to a potential loss of identity information.

---

[1] https://github.com/leondgarse/Keras_insightface
[2] https://github.com/TencentARC/GFPGAN
[3] https://github.com/yangxy/GPEN
[4] https://github.com/TencentYoutuResearch/FaceRestoration-sgpn

TABLE I
SIMILARITY MEDIANS

| model | GFP-GAN [20] | GPEN [21] | SGPN [26] |
|---|---|---|---|
| AdaFace [29] | 0.565 | 0.523 | 0.406 |
| MagFace [30] | 0.518 | 0.527 | 0.438 |
| ArcFace [31] | 0.672 | 0.649 | 0.587 |

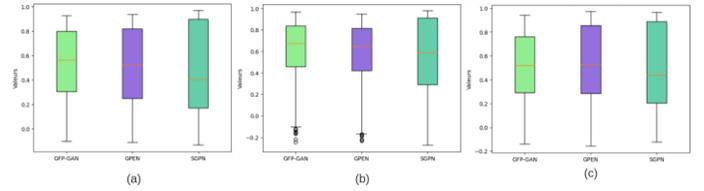

Fig. 4. Boxplots of similarity index between embeddings extracted using (a)AdaFace, (b)MagFace and (c)ArcFace from LFW images and restored images using SGPN, GFP-GAN and GPEN face restoration techniques

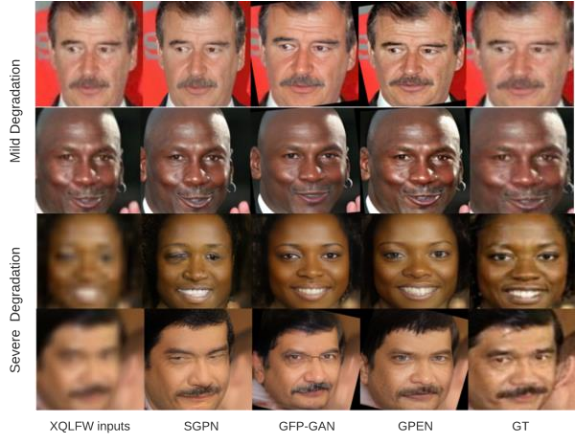

Fig. 3. Perceptual comparison between restored images using SGPN, GFP-GAN and GPEN face restoration techniques

*2) Identity Preserving Quantification:* To quantify the potential of face restoration techniques in preserving the identity information within the enhanced images, the statistics of the cosine similarity between the restored XQLFW and the original LFW images are analyzed. We calculate the cosine similarity between the embeddings of the two images extracted using AdaFace based on the ResNet100 [?] backbone and trained on WebFace4m [3] database, MagFace based on ResNet50 backbones and trained on MS1MV2 [31] and ArcFace with ResNet50 backbone trained on the CASIA-WebFace [?] database. These statistics are represented via boxplots in Fig. 4. It is easy to notice that for the three face embeddings, SGPN shows a high interquartile range or variability. The other BFR techniques exhibits lower interquartile range with higher medians. The **Cosine Smilarity Median (CSM)** can be considered as a good metric for identity preservation. Table I summarizes the obtained CSM values and we can advocate that the GFP-GAN technique possess the highest identity preserving capability when used in conjunction AdaFace and ArcFace. Meanwhile, for the MagFace model, images restored using GPEN demonstrate the highest similarity to the ground truth images. The lower CSM's quantities obtained using SGPN confirm its key pitfall in preserving the valuable identity information, especially for severely degraded face images.

*3) Biometric Utility Quality via MagFace:* In order to shed light on the impact of the studied BFR techniques concerning the biometric utility quality metrics, which are highly correlated to the recognition accuracy, the Magface metric, which simply measures the magnitude of the MagFace embedding, is employed. The magnitude of the face features vectors extracted using MagFace from the original XQLFW and its restored versions using GFP-GAN, GPEN and SGPN. The distribution plots depicting these magnitudes are presented in Fig. 5. For the XQLFW dataset, we observe a right-skewed distribution, indicating that the majority of images are located on the left side of the graph and have low magnitudes, thus indicating low quality. Regarding the GPEN and GFP-GAN graphs, we can observe a near-Normal distribution. The peak of the distribution for GPEN is centered on 19, while for GFP-GAN, it is centered on 18. These distributions suggest that there is a predominance of high- quality images. On the other hand, for SGPN, although there is a decrease in the number of low-quality images, the frequency of the high-quality images does not reach the same level as that of GPEN and GFP-GAN. This suggests that SGPN may have a comparatively lower performance compared to the other two BFR techniques in terms of improving the biometric quality of the sample.

*4) Effect on face verification performance:* Previous findings support the hypothesis that GFP-GAN and, to a lesser extent, GPEN could be good candidates for boosting the face verification performance of low-quality images. Before investigating the verification accuracy after applying face restoration, we run our first experiments to see the performance drop when passing from LFW to XQLFW. The results depicted in table II show a significant drop in performance by around 12% for AdaFace, 16% and 20% for MagFace trained on ResNet100 and ResNet50 respectively, and even reaching 26% for ArcFace. We can observe that AdaFace is the most robust to degradation and most suited to handle low quality face verification. The drastic drop in performance across different face recognition models serves as compelling evidence for the importance of addressing image degradations. To this end, the best performing configuration for each loss function, namely AdaFace trained on webface4m, MagFace with ResNet50 backbone and ArcFace with ResNet50 backbone are selected for our last experiments to evaluate the impact of blind face restoration on verification accuracy. The results are presented in table III.

The obtained results highlight the outperformance of GFP-GAN compared to the other BFR techniques. A significant

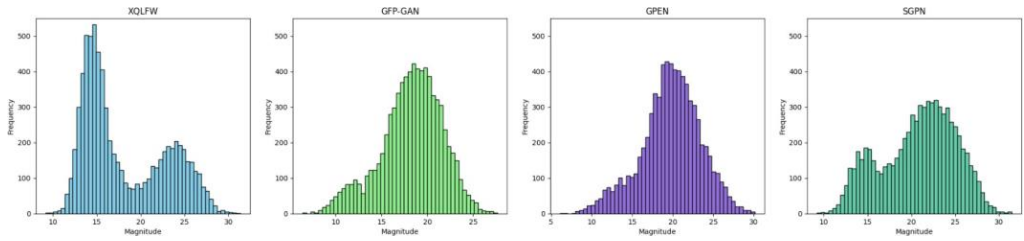

Fig. 5. Statistical distributions of the Magface quality metric on the original XQLFW, and on the restored XQLFW using GFP-GAN, GPEN and SGPN respectively.

TABLE II
FACE VERIFICATION ACCURACIES OF SOTA SYSTEMS ON BOTH LFW AND XQLFW

| Cost function | Backbone architecture | Training dataset | accuracy | |
|---|---|---|---|---|
| | | | LFW | XQLFW |
| AdaFace [29] | ResNet100 | WebFace4m | **0.998** | **0.879** |
| MagFace [30] | ResNet100 | MS1MV3 | 0.996 | 0.806 |
| | ResNet500 | MS1MV2 | 0.998 | 0.835 |
| ArcFace [31] | ResNet50 | CASIA-WebFace | 0.993 | 0.737 |

Values in bold indicate the best accuracy for each experiment.

TABLE III
FACE VERIFICATION ACCURACIES ON THE XQLFW DATABASE AFTER BLIND FACE RESTORATION

| model | bef/rest | SGPN [26] | GFP-GAN [20] | GPEN [21] |
|---|---|---|---|---|
| AdaFace [29] | 0.879 | 0.817 | 0.896 | 0.857 |
| MagFace [30] | 0.835 | 0.791 | 0.863 | 0.836 |
| ArcFace [31] | 0.737 | 0.761 | 0.865 | 0.82 |

gain in performance is achieved across all the three models, with an absolute increase in accuracy ranging from 1.7% to 12.8%. However, it is important to note that the performance of the other models can exhibit both improvements and declines. Specifically, when evaluating GPEN, the performance on ArcFace demonstrates a significant 9% increase. There is a marginal improvement of 0.1% for MagFace, while there is a decrease of 2% for AdaFace. On the other hand, when evaluating SGPN, there is a decline in performance ranging from 4% to 6%. These unexpected findings suggest that the restoration models, while enhancing perceptual quality, may inadvertently cause a loss of identity-related information, resulting in a decrease in performance. It is important to mention that the obtained recognition results are perfectly inline with our analysis concerning similarity index and biometric quality.

## V. CONCLUSION

In this study, we examined the effectiveness of face restoration techniques based on gen- erative priors in preserving face identity information. Our focus was on their impact on improving the performance of face recognition models, specifically in the verification task. Our findings revealed that the AdaFace technique outperformed other methods in this task across images of varying qualities. Moreover, the GFP-GAN restoration technique excelled in enhancing visual quality and preserving identity information, enabling accurate verification of facial recognition patterns. To leverage these findings, we proposed a novel face verification system that integrated a face restoration module utilizing the GFP-GAN technique at the early stage of the verification process. This system aimed to enhance the face verification performance of the AdaFace, MagFace, and ArcFace face recognition techniques in challenging environments with complex image distortions. While significant performance improvements were achieved in the face verification task using the XQLFW dataset with synthetic degradations [32], it is crucial to conduct additional testing on real-world image databases to validate the effectiveness of the proposed solution in practical scenarios. Furthermore, it is imperative to carry out further research to gain valuable insights into the practical implications of employing blind face restoration methods in the field of face recognition.